\documentclass[journal]{IEEEtai}

\usepackage[colorlinks,urlcolor=blue,linkcolor=blue,citecolor=blue]{hyperref}

\usepackage{color,array}

\usepackage{graphicx}
\usepackage{cite}
\usepackage{amsmath}
\usepackage{amssymb,amsfonts}
\usepackage{graphicx}
\usepackage{textcomp}
\usepackage{xcolor}
\usepackage{multirow}
\usepackage{algorithm}
\usepackage{algpseudocode}
\usepackage{algorithmicx}
\usepackage{pifont}
\begin{document}

\title{Dynamic Long-Term Time-Series Forecasting via Meta Transformer Networks}

\author{Muhammad Anwar Ma'sum,
        MD Rasel Sarkar,
        Mahardhika Pratama,~\IEEEmembership{Senior Member,~IEEE}, Savitha Ramasamy ~\IEEEmembership{Senior Member, ~IEEE}, Sreenatha Anavatti, Lin Liu, Habibullah, Ryszard Kowalczyk% <-this % stops a space

\thanks{M. A. Ma'sum, M. Pratama, L. Liu, Habibullah and R. Kowalczyk are with STEM, University of South Australia, Adelaide,
SA, 5095 Australia. MD. R. Sarkar, S. Anavatti are with SEIT, University of New South Australia, Canberra, Australia. S. Ramasamy is with I2R, A*Star, Singapore. R. Kowalczyk is also with System Research Institute, Polish Academy of Science, Poland.}
\thanks{M. A. Ma'sum, MD. R. Sarkar, M. Pratama share equal contributions}
\thanks{M. Pratama is the corresponding author.}}

\markboth{Journal of IEEE Transactions on Artificial Intelligence, Vol. 00, No. 0, Month 2020}
{First A. Author \MakeLowercase{\textit{et al.}}: Bare Demo of IEEEtai.cls for IEEE Journals of IEEE Transactions on Artificial Intelligence}

\maketitle

\begin{abstract}
A reliable long-term time-series forecaster is highly demanded in practice but comes across many challenges such as low computational and memory footprints as well as robustness against dynamic learning environments. This paper proposes Meta-Transformer Networks (MANTRA) to deal with the dynamic long-term time-series forecasting tasks. MANTRA relies on the concept of fast and slow learners where a collection of fast learners learns different aspects of data distributions while adapting quickly to changes. A slow learner tailors suitable representations to fast learners. Fast adaptations to dynamic environments are achieved using the universal representation transformer layers producing task-adapted representations with a small number of parameters.  Our experiments using four datasets with different prediction lengths demonstrate the advantage of our approach with at least $3\%$ improvements over the baseline algorithms for both multivariate and univariate settings. Source codes of MANTRA are publicly available in \url{https://github.com/anwarmaxsum/MANTRA}. 
\end{abstract}

\begin{IEEEImpStatement}
Time-series forecasting plays a vital role in many application domains including but not limited to health, manufacturing, finance, etc. The advent of the transformer has driven significant progress in time-series forecasting but incurs prohibitive complexity when dealing with long-term forecasting involving long sequences. Recent advances have attempted to address the long-term time-series forecasting problems but overlook the problem of dynamic environments which might affect time-series patterns. Our work aims to tackle the problem of dynamic long-term time-series forecasting where the proposal of MANTRA is put forward. Our numerical studies find the advantage of MANTRA over prior arts where it delivers improved accuracy with noticeable margins.  
\end{IEEEImpStatement}


\section{Introduction}
\IEEEPARstart{T}{ime} series forecasting problems play a vital role in many applications where it aims to predict future values given a sequence of current and past observations. This problem becomes even more challenging than that in the long-term setting where a model has to handle long sequences. In addition, a time-series forecasting problem is inherent with non-stationary behaviors confirming the case of dynamic long-term time-series forecasting problem, i.e., the offline time-series forecaster trained in the offline fashion is quickly outdated and requires a retraining process from scratch if a new data pattern is observed. In a nutshell, the long-term time-series forecasting problem calls for a machine learning algorithm to adapt quickly to changes with low computational and memory overheads. 

The inception of deep learning technologies have revolutionized the field of machine learning including the time-series forecasting problems. Deep learning approaches include an automatic feature engineering step exploring hidden patterns of data samples, thereby resulting in improved generalization powers. Nevertheless, it is generally accepted that the use of deep learning strategies in the dynamic cases remain challenging. Recent research efforts \cite{Zhou2012OnlineIF,Sahoo2018OnlineDL,Das2019MUSERNNAM,Pratama2019AutomaticCO,Ashfahani2019AutonomousDL} have been devoted to study the case of dynamic environments but all of them are designed for classification rather than regression. In addition, these approaches are not designed for long-term time-series forecasting cases involving long predictive horizons. 

An ensemble approach \cite{Lughofer2021OnlineBO,Pratama2018EvolvingEF,Pratama2020AnIC,Cerqueira2017ArbitratedEF,Saadallah2021ExplainableOD} is one of the most proven techniques for dealing with non-stationary environments because it allows each base learner to master different regions of data space and handles the bias-variance problem better than the single model approach. The influence of base learners can be controlled in such a way as to reflect the current concept. The key lies in the online model selection evaluating each base learner for inferences. Most ensemble approaches are still trained in the traditional manner without fully exploiting the representational learning power of deep learning technologies. \cite{Eldele2021TimeSeriesRL} proposes a transformer approach with the self-supervised or contrastive learning approach making it possible to extract meaningful representations for downstream tasks. It also offers suitable augmentation strategies of contrastive learning usually applied for visual data for time series data. The transformer offers some potential for time-series forecasting tasks compared to RNNs and LSTMs \cite{Zhou2021InformerBE} where it offers a simpler structure than those the two approaches and ignores the recurrent structure. However, the contrastive learning strategy is slow and calls for large mini-batch sizes. The transformer approach also does not work well for long-term time-series forecasting problems because of the high computational and memory costs of full attention mechanisms \cite{Wu2021AutoformerDT}. A sparse version of the self-attention module is used to scale well in the long-term context \cite{Zhou2021InformerBE} but suffers from reduced information. \cite{Wu2021AutoformerDT} proposes the autoformer model going beyond the transformer-based approach where the key lies in the concept of auto-correlation and series decomposition. Despite its power for time-series forecasting, this approach has not been investigated for dynamic settings. Generally speaking, the application of deep learning approaches for dynamic long-term time-series forecasting is scarce and deserves an in-depth study.  

Meta-transformer network (MANTRA) is proposed in this paper for dynamic long-term time-series forecasting problems. MANTRA proposes the concept of the extended dual networks \cite{Pham2021DualNetCL} putting forward a slow learner and an array of fast learners working cooperatively. The slow learner functions to produce useful representations for downstream tasks via the self-supervised learning procedure while an ensemble of fast learners are deployed to explore different regions of data space and to adapt quickly to changing data distributions. That is, the slow training procedure of the self-supervised training approach is addressed by decoupling the array of the fast learners from the slow learner. The issue of drifting distributions is overcome by the proposal of a universal representation transformer (URT) concept producing task-adapted representations with few parameters \cite{Liu2021AUR} assuring fast adaptations. The original URT layer is generalized for dynamic time-series forecasting problems rather than image classification problems and for task-adapted representations of the ensemble of fast learners. MANTRA is based on the autoformer \cite{Wu2021AutoformerDT} approach in this paper putting forward the auto-correlation and series decomposition modules structured under the encoder-decoder configuration. The series decomposition module enables separation between trend-cyclical and seasonal parts pinpointing long-term dependencies. Nonetheless, the concepts of MANTRA can be generalized to other network structures such as Informer \cite{Zhou2021InformerBE}, FEDformer \cite{Zhou2022FEDformerFE}. It also fits well for recent findings \cite{Zeng2022AreTE} showing the advantage of linear modules over transformer approaches.  

This paper offers three major contributions: 
\begin{enumerate}
    \item the concept of extended dual networks is proposed for dynamic long-term time-series forecasting problems. Ours differ from \cite{Pham2021DualNetCL} due to the use of autoformer rather than convolutional learners and the problem of dynamic time-series forecasting. We also replace the slow learner objective minimizing the controlled reconstruction loss \cite{Chowdhury2022TARNetTR} allowing less computations and less sensitive to augmentations than that of the Barlow Twins approach \cite{Zbontar2021BarlowTS}. In addition, the concept of extended dual networks fits an ensemble of fast learners rather than a single fast learner;
    \item the URT concept is proposed here for the dynamic long-term time-series forecasting problems to adapt quickly to the concept changes. Our approach distinguishes itself from \cite{Liu2021AUR} where the URT concept dynamically aggregates the final output of the array of fast learners rather than the backbone networks for few-shot learning;
    \item the source codes of MANTRA are already made publicly available in \url{https://github.com/anwarmaxsum/MANTRA}.
\end{enumerate}
Our numerical results show that MANTRA delivers the most encouraging results compared to baseline algorithms with at least $3\%$ margins. 

\section{Related Works}
\subsection{Time-Series Forecasting}
Time-series forecasting problems are vital for many real-world applications and have been an active research topic. One early approach for time-series forecasting problems is ARIMA \cite{Box1970TimeSA} converting the non-stationary components into the stationary components. Another popular approach is via a filtering approach \cite{Bzenac2020NormalizingKF}. The advent of deep learning technologies has driven the time-series forecasting field. Recurrent Neural Networks (RNNs) are applied to model the temporal patterns of time-series data \cite{Wen2017AMQ,Rangapuram2018DeepSS,Flunkert2017DeepARPF}. The skipped connection is proposed in convolutional neural networks \cite{Lai2018ModelingLA} to capture the short-term and long-term temporal characteristics. The idea of attention is incorporated for time-series forecasting problems to unveil the long-term dependencies of time-series data \cite{Song2018AttendAD,Qin2017ADA} while other works rely on the causal convolution \cite{Borovykh2017ConditionalTS,Bai2018AnEE}.

Transformer \cite{Vaswani2017AttentionIA} has demonstrated excellent performances for sequential data. The bottleneck of transformer for time-series forecasting problems lies in the self-attention mechanism becoming prohibitive in the long-term time-series forecasting cases. Recent works overcome this issue with the proposal of sparse self-attention improving its scalability for such problems \cite{LI2019EnhancingTL,Zhou2021InformerBE,Kitaev2020ReformerTE}. These approaches remain a vanilla transformer depending on the point-wise dependency and aggregation. In \cite{Wu2021AutoformerDT}, the concept of autoformer is designed. It goes beyond the transformer structure where long-term dependencies of time-series data are explored using series decomposition and auto-correlation modules. Recently, \cite{Zhou2022FEDformerFE} developed an extension of Autoformer where the concept of Fourier transform is incorporated. Recent finding \cite{Zeng2022AreTE} demonstrates that embarrassingly simple models beat the transformer-based approaches for long-term time-series forecasting. Nonetheless, all of these approaches still cover the static cases where time-series patterns are assumed to be unchanged. Our approach is also structure-independent where it can fit with any base models for long-term time-series forecasting problems.  

\subsection{Deep Learning for Dynamic Environments}
The deep learning technologies are known to be slow because of their iterative nature. \cite{Zhou2012OnlineIF} proposes an online incremental feature learning featuring a series of generative steps minimizing the reconstruction loss and the discriminative step minimizing the cross-entropy loss. It also puts forward a network-growing procedure based on the reconstruction loss of the generative phase. The concept of the hedge back-propagation and the different-depth network structure was proposed in \cite{Sahoo2018OnlineDL} for online classification problems. 

\cite{Ashfahani2019AutonomousDL} proposes a flexible deep neural network having an elastic network width and depth. It is based on the different-depth network structure where every layer produces its local output and the final output is aggregated using the weighted voting mechanism. The same principle is applied in \cite{Pratama2019AutomaticCO} under the framework of a multi-layer perceptron (MLP) structure. The concept of adaptive memory and soft-forgetting is proposed. A recurrent neural network based on the teacher-forcing principle is proposed in \cite{Das2019MUSERNNAM}. These works are designed only for classification problems rather than for regression problems. In addition, they are not designed for any forecasting problems. 

The concept of deep neural networks have been utilized for time-series forecasting in \cite{Cerqueira2017ArbitratedEF} where a meta-learning approach is adopted to aggregate the output of ensemble classifiers. The concept of saliency map is applied in \cite{Saadallah2021ExplainableOD} for an online model selection of a deep ensemble classifier. The meta-learning strategy based on MAML \cite{Finn2017ModelAgnosticMF} is proposed in \cite{You2021LearningTL} to adapt quickly in the presence of concept drift. The use of deep learning technologies for dynamic long-term time-series forecasting is under-explored due to two reasons: 1) \cite{Cerqueira2017ArbitratedEF,Saadallah2021ExplainableOD,You2021LearningTL} rely on a traditional learning technique based on the MSE loss which does not yet explore the representational power of deep neural networks. The application of contrastive learning for time-series forecasting has been championed in \cite{Eldele2021TimeSeriesRL} but this approach does not scale well for dynamic settings because the contrastive learning usually calls for a large mini-batch size. MANTRA addresses this flaw with a decoupling strategy between an ensemble of fast learners and a slow learner; 2) our strategy to deal with the concept drift problems is based on the URT concept \cite{Liu2021AUR} generalized here to address the dynamic time-series forecasting problems and to create task-adapted representations of the array of the fast learners. Such an approach adapts quickly to concept drifts because backbone networks can be frozen while only tuning relatively small numbers of the URT parameters. 
\section{Preliminary}
\subsection{Problem Formulation}
Given a time-series $X$ with a length of $T$, $X=[x_1,x_2,...,x_T]$, where $x_t$ is a component of time-series at a $t-th$ time instant, the goal of time-series forecasting problem is to perform step-ahead predictions $t\geq T$ using a model $f$ parameterized with $\theta$ where the model maps the input space to the target space $f_{\theta}: \mathcal{X} \rightarrow \mathcal{Y}$ and \textcolor{red}{$x_t\in\Re^{D}$ and $y_t\in\Re$ are the input and target of the model.} $D$ is the number of features that forms a single observation $x_t$ at the $t-th$ time index and $f_{\theta}(.)$ is represented by the autoformer here \cite{Wu2021AutoformerDT} but applicable to other structures. The target attribute  $y_t=x_{t+\tau}$ is inherent to future values of the time-series input $x_t$ where $\tau$ is the number of time-step. Long-term time-series forecasting is considered here where a well-ahead prediction is made or $\tau$ is large. A fully supervised time-series forecasting problem is considered here where $(x_t,y_t)\in \mathcal{X}\times\mathcal{Y}$. The underlying challenge lies in the concept drift problem \cite{Gama2014ASO} presenting the case of changing data distributions $P(X,Y)_t\neq P(X,Y)_{t+1}$ where the speeds and magnitudes of the concept drift are hidden to the model $f_{\theta}(.)$. The model $f_{\theta}(.)$ is supposed to detect the concept drifts in timely fashions and to recover quickly from them. The training process is governed by a minimization of a predictive loss function $\arg\min_{\theta}\mathcal{L}$ where $\mathcal{L}=-\log P_{\theta}(Y_{1:T}|X_{1:T})=\sum_{t=1}^{T}(y_t-f_{\theta}(x_t))^2$.

\subsection{Autoformer}
MANTRA borrows the Autoformer concept \cite{Wu2021AutoformerDT} to play roles as the slow learner $g_{\psi}(.)$ as well as the fast learners $\{f_{\theta_{i}}(.)\}_{i=1}^{M}$. The Autoformer features an encoder-decoder structure having the auto-correlation block and the series-decomposition block. The series decomposition block separates the cyclical part from the seasonal part using the moving average to capture periodic fluctuations and find the long-term pattern \cite{Wu2021AutoformerDT}. The moving average is implemented as the average pooling operator with padding to assure consistent time-series length. The auto-correlation block unveils period-based dependencies by means of calculation of auto-correlation and combines similar sub-series with time-delay aggregations. Similar sub-series are those in the same phase, i.e., the series are rolled according to selected time-delay. This mechanism is claimed to replace the self-attention mechanism seamlessly. The period-based dependencies are computed with the Fast Fourier Transform (FFT) method and combined with the output of the period-based aggregation module to obtain the auto-correlations.

The series decomposition block is designed to separate the series into trend-cyclical components and seasonal components \cite{Wu2021AutoformerDT} pinpointing the long-term progressions and seasonality of the series. This is done by applying the moving average to remove periodic fluctuations and highlight the long-term trends. For series with the length of $L$ $X\in\Re^{L\times D}$, the series decomposition block is expressed:
\begin{equation}
    \begin{split}
        X_{cy}=AvgPool(Padding(X))\\
        X_{se}=X-X_{cy}
    \end{split}
\end{equation}
where $X_{se},X_{cy}\in\Re^{L\times D}$ denote the seasonal and trend-cyclical parts respectively. $AvgPool(.)$ with the padding operation is implemented for the moving average such that the length of the series remains unchanged. The series decomposition block is formalized as $X_{se},X_{cy}=SD(X)$.

The auto-correlation module unveils the period-based dependencies by computing the series autocorrelation and aggregates similar sub-series by time delay aggregation. The time delay aggregation block rolls the series based on selected time delay $\tau_1,...,\tau_k$, thus aligning similar sub-series that are at the same phase position of estimated periods. For a single-head situation, after projection, we obtain query $Q$, key $K$, and value $V$, and the auto-correlation module replaces the self-attention mechanism seamlessly.

\begin{equation}
    \begin{split}
        \tau_1,...,\tau_k=\arg Topk_{\tau\in\{1,...,L\}} (R_{Q,K}(\tau))\\ \hat{R}_{Q,K}(\tau_1),...,\hat{R}_{Q,K}(\tau_k)=SoftMax(R_{Q,K}(\tau_1),...,R_{Q,K}(\tau_k))\\AutoCorrelation(Q,K,V)=\sum_{i=1}^{k}Roll(V,\tau_i)\hat{R}_{Q,K}(\tau_i)
    \end{split}
\end{equation}
where $\arg Topk(.)$ returns the top k autocorrelations \textcolor{red}{indices} while $R_{Q,K}$ denotes the autocorrelation between series $Q$ and $K$. $Roll(X,\tau)$ stands for the shifting operation to $X$ with time delay $\tau$. The multi-head version is realized by concatenating all heads and in turn producing the final output by projection via $W_{out}$ where each head is represented by the auto-correlation.

The encoder focuses on the seasonal characteristics of the time-series data and removes the long-term cyclical characteristics while the decoder combines the extracted information of the encoder and other information, i.e., the seasonal part and the trend-cyclical part, to deliver the final predictions. 
The encoder receives a mini-batch of time-series data containing the past $B$ records $X_{B}\in\Re^{B\times D}$ while the input of decoder is both the seasonal part $X_{s}\in\Re^{(\frac{B}{2}+O)\times D}$ and trend-cyclical part $X_{c}\in\Re^{(\frac{B}{2}+O)\times D}$ produced by the series decomposition block. Note that $O$ is simply a placeholder filled with zero. The second half of the data batch is fed to the decoder to provide recent information.
\begin{equation}
    \begin{split}
        X_{se}, X_{cy} = SD(X_{\frac{B}{2}:B})\\
        X_{s} = concat(X_{se},X_{0})\\
        X_{c} = concat(X_{cy},X_{mean})
    \end{split}
\end{equation}
where $X_{se},X_{cy}\in\Re^{\frac{B}{2}\times D}$ denote the output of the series decomposition block $SD(.)$ presenting the seasonal information and the trend-cyclical information. $X_0,X_{mean}$ are the placeholder with zero entries and the mean of $X_{B}$ respectively. 

A single encoder $h_{enc}^{l}=Enc(h_{enc}^{l-1})$ is composed of stacked auto-correlation, series-decomposition, and feed-forward modules:
\begin{equation}
    \begin{split}
        S_{en}^{l,1},\_ = SD(AC(h_{enc}^{l-1})+h_{enc}^{l-1})\\
        S_{en}^{l,2},\_ = SD(FF(S_{en}^{l,1})+S_{en}^{l,1})
    \end{split}
\end{equation}
where $h_{enc}^{l}=S_{en}^{l,2}$ is the output of the $l-th$ encoder, and $\_$ stands for the eliminated long-term cyclical trend of the series decomposition module. That is, the encoder only focuses on the seasonal part of the time-series data. %Here, we exemplify the case of fast learner $h_{enc}^{l}$ and the same steps are applied for the slow learner $u_{enc}^{l}$. 

\begin{figure*}
    \centering
    \includegraphics[scale=0.35]{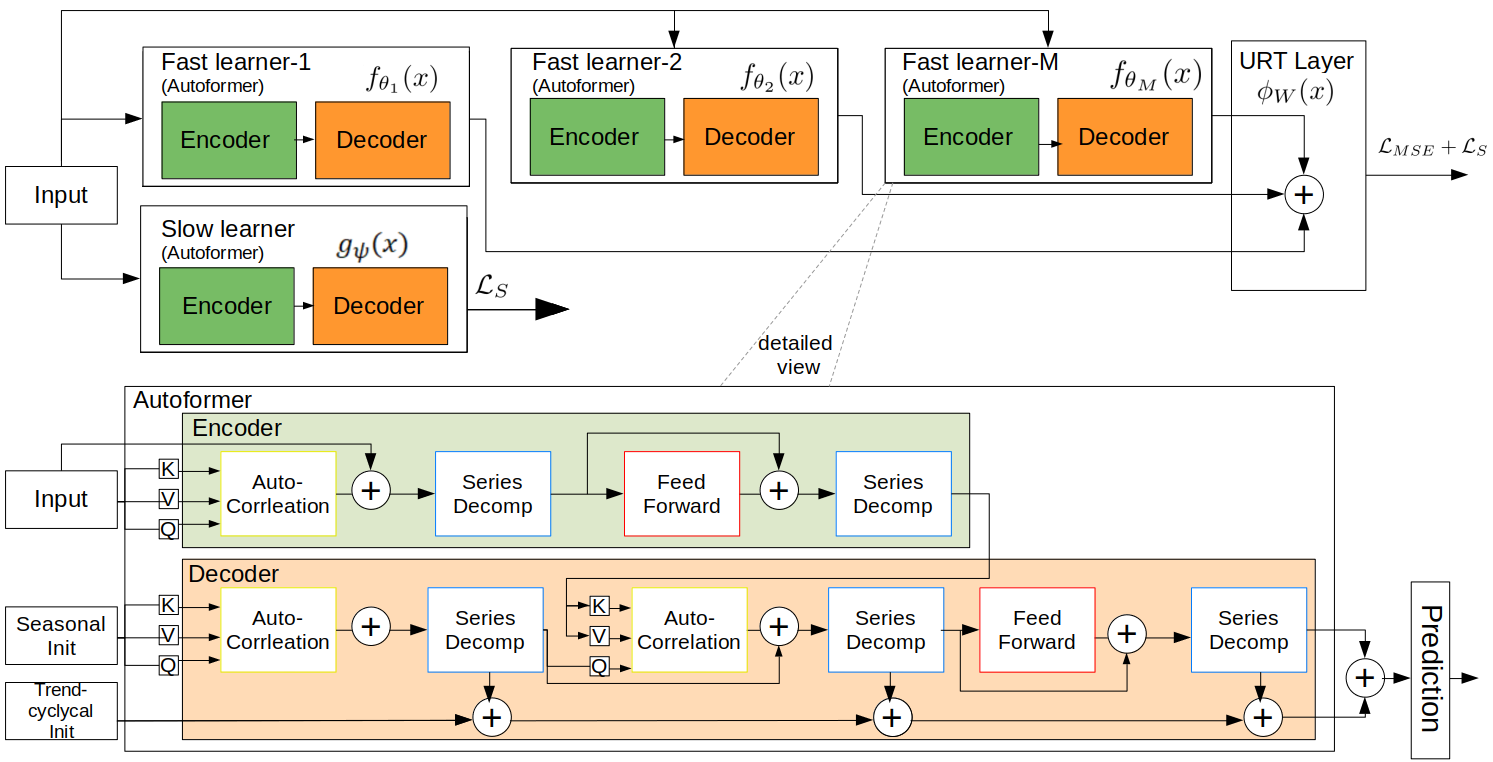}
    \caption{MANTRA is consolidated by an ensemble of fast learners and a slow learner where the final output is aggregated by a URT layer. Fast learners and URT layer are trained in the end-to-end fashion minimizing the mean-squared error loss while slow learner is learned via the self-supervised learning loss, controlled reconstruction loss.}
    \label{fig:MANTRA}
\end{figure*}

As with the encoder, a decoder is configured as a stack of the auto-correlation module, the series decomposition module, and the feed-forward module. Suppose that there exist $L$ decoders, the $l-th$ decoder $h_{dec}^{l}=Dec(h_{dec}^{l-1},h_{enc}^{L})$ is expressed:
\begin{equation}
    \begin{split}
        S_{de}^{l,1},C_{de}^{l,1}=SD(AC(h_{de}^{l-1})+h_{de}^{l-1})\\
        S_{de}^{l,2},C_{de}^{l,2}=SD(AC(S_{de}^{l,1},h_{en}^{L})+S_{de}^{l,1})\\
        S_{de}^{l,3},C_{de}^{l,3}=SD(FF(S_{de}^{l,2})+S_{de}^{l,2})\\
        C_{de}^{l}=C_{de}^{l-1}+\theta_{l,1}C_{de}^{l,1}+\theta_{l,2}C_{de}^{l,2}+\theta_{l,3}C_{de}^{l,3}
    \end{split}
\end{equation}
where $S_{de}^{l,i},C_{de}^{l,i},i\in\{1,2,3\}$ denote the seasonal and trend-cyclical components after applying the $i-th$ series decomposition block $SD(.)$. $\theta_{l,i},i\in\{1,2,3\}$ is a projection vector of the trend-cyclical components. Given $L$ decoders, the output of autoformer is inferred by a weighted sum operation between the seasonal and trend-cyclical components $f_{\theta}(x)=\theta_S S_{de}^{l,3}+C_{de}^{L},g_{\psi}(x)=\psi_S S_{de}^{l,3}+C_{de}^{L}$ where $\theta_S,\psi_S$ are the weight vectors to project the seasonal component into the same dimension as the output. The slow learner follows exactly the same steps as the fast learner. \textcolor{red}{Fig. \ref{fig:MANTRA} exhibits the architecture of MANTRA. It comprises an ensemble of fast learners and a slow learner trained with different objectives. The features of both learners are combined before delivering the final predictions.}

 %The modulation of the fast learner and the slow learner occurs in every layer of the encoder-decoder structure of the Autoformer. That is, it is implemented at the output of encoder $h_{enc,l}^{i}=u_{enc,l}\otimes\hat{h}_{enc,l}^{i}$ since the slow learner is built upon the encoder of Autoformer rather than as the full encoder-decoder architecture. Note that the goal of slow learner is to extract meaningful representations instead of to generate predictions.
\section{Meta-Transformer Networks (MANTRA)}
\subsection{Fast and Slow Learning Approach}
MANTRA is developed from the concept of the extended dual-Net where an array of fast learners and a slow learner work cooperatively to address the dynamic time-series forecasting problems. This approach addresses the slow characteristic of contrastive learning because the slow learning process is decoupled from the fast learning process. The goal of the slow learner is to extract meaningful or general representations for downstream tasks while the fast learner adapts quickly to new patterns and combines general representations supplied by the slow learners. Our innovation here is to extend the concept of DualNet for the dynamic time-series forecasting problem having different characteristics of those continual image classification problems and applies the controlled time-series reconstruction strategy \cite{Chowdhury2022TARNetTR} as the slow learning approach rather than the Barlow Twins (BT) strategy \cite{Zbontar2021BarlowTS}. The controlled reconstruction strategy is preferred over the BT approach because it incurs fewer computations than the BT approach and is independent of any augmentations. In addition, an ensemble of fast learners are deployed here rather than a single fast learner handling the bias-variance tradeoff and the concept drifts better than a single fast learner. 

The slow learner $g_{\psi}(.)$ is trained to minimize the masked loss function $\mathcal{L}_{m}$ without any labeled samples \cite{Chowdhury2022TARNetTR}. The controlled masking strategy is implemented where $m_t$ is a binary mask at the $t-th$ time step. A $D$ dimensional time-series input $\tilde{x}_t$ is zeroed if $m_t=1$. $\mathcal{L}_m$ is realized as the mean-squared error (MSE) loss function and applied to both masked and unmasked time-series samples with the original input $x_t$ as the target variable. 
\begin{equation}
    \mathcal{L}_{m}=\frac{1}{D\sum_{t=1}^{T}m_t}\sum_{t=1}^{T}m_t||g_{\psi}(\tilde{x}_t)-x_t||_2^{2}
\end{equation}
\begin{equation}
    \mathcal{L}_{um}=\frac{1}{D(S-\sum_{t=1}^{T}m_t)}\sum_{t=1}^{T}(1-m_t)||g_{\psi}(\tilde{x}_t)-x_t||_2^{2}
\end{equation}
Both $\mathcal{L}_{m}$ and $\mathcal{L}_{um}$ are combined with a tradeoff parameter $0\leq\lambda\leq1$ to train the slow learner rather than only the masked components as per \cite{Zerveas2021ATF} since the reconstruction process of the masked input heavily relies on the reconstruction process of the unmasked input. The overall loss function is formalized:
\begin{equation}
    \mathcal{L}_{S}=\lambda\mathcal{L}_{m}+(1-\lambda)\mathcal{L}_{um}
\end{equation}
The controlled reconstruction process selects important time-stamps $t$ \cite{Chowdhury2022TARNetTR} rather than relying on random selections \cite{Zerveas2021ATF}. The selection strategy is done by checking the attention matrix \cite{Chowdhury2022TARNetTR} which measures the matching degree between the keys and the queries. This is done differently here where important time steps are determined across the top k autocorrelations $\hat{\mathcal{R}}_{\mathcal{Q},\mathcal{K}}\in\Re^{k\times k}$ because of the use of autoformer and the long-term time-series problem involving long sequences. Note that the top k autocorrelations have been extracted in AutoFormer. This strategy risks the selection of the same subset at every epoch, i.e., the over-fitting problem. The regularization strategy \cite{Chowdhury2022TARNetTR} is applied to alleviate this problem.

An array of fast learners $M$ are deployed here and created by applying different initialization strategies to capture different aspects of data distributions where $M$ stands for the array size. Suppose that $\{f_{\theta_i}(.)\}_{i=1}^{M}$ denotes the $i-th$ fast learner and $g_{\psi}(.)$ labels the slow learner, $S_{de,\theta}^{l,i},S_{de,\psi}^{i},C_{de,\theta}^{l,i},C_{de,\psi}^{i}$ stand for seasonal and cyclical components of the $l-th$ fast learner and the slow learner. Decoder features of the slow learner and the $l-th$ fast learners are concatenated before being projected back to the output dimension \cite{Pham2021DualNetCL,Perez2018FiLMVR}:
\begin{equation}
    \hat{S}_{de,\theta}^{l,3}=S_{de,\theta}^{l,3}\oplus S_{de,\psi}^{3};\quad \hat{C}_{de,\theta}^{l,3}=C_{de,\theta}^{l,3}\oplus C_{de,\psi}^{3}
\end{equation}
where $\oplus$ denotes the concatenation operation which occurs at the last layer of Autoformer. The final output of the $l-th$ fast learner is inferred as $f_{\theta}(x)=\theta_S \hat{S}_{de,\theta}^{l,3}+\theta_{C}\hat{C}_{de,\theta}^{l,3}$ where $\theta_{S},\theta_{C}$ are projectors to modify the output dimension. A mean squared error (MSE) loss function $\mathcal{L}_{MSE}$ is applied to train all fast learners as follows:
\begin{equation}\label{MSEloss}
\mathcal{L}_{MSE}=\sum_{t=1}^{T}\frac{1}{2}(y_t - \phi_{W}(\oplus(\{f_{\theta_i}(x_t)\}_{i=1}^M)))^2    
\end{equation}
where $\phi_{W}(\oplus(\{f_{\theta_i}(x)\}_{i=1}^{M}))$ stands for the final output of MANTRA and $y_t$ denotes the target variable at the $t-th$ time instant. Note that the case of a single-head URT layer is presented in $\phi_{W}(\oplus(\{f_{\theta_i}(x)\}_{i=1}^{M}))$ while a generalized version, a multi-head, is elaborated in the next section. Nonetheless, our experiment shows that the use of only the MSE loss function is not sufficient to fully exploit the dual net structure. Hence, all fast learners are trained minimizing the joint loss function $\mathcal{L}_{MSE}+\mathcal{L}_{S}$. 

\subsection{Drift Handling Mechanism via the URT Layer}
MANTRA makes use of an ensemble of fast learners with $M$ components connected to a single-head URT layer having a single attention head delivering the final predictions. An array of fast learners $\{f_{\theta_i}(.)\}_{i=1}^{M}$ are trained alongside with the URT layer $\phi_{W}(.)$ minimizing \eqref{MSEloss}. MANTRA is sufficiently flexible where if the $M$ fast learners $\{f_{\theta_i}(.)\}_{i=1}^{M}$ are pre-trained using different but related problems, e.g., different domains \cite{Liu2021AUR}, one can fix the fast learners and train the URT layer $\phi_{W}(.)$ only. The URT concept is inspired by the self-attention mechanism of transformer \cite{Vaswani2017AttentionIA} deriving the attention mechanism via the query-key pairs. Since the original URT layer \cite{Liu2021AUR} is designed for few-shot image classification problems, we offer some modifications here. The URT layer concatenates the representations of the $M$ fast learners and converts them into a final output taking into considerations of attentions:
\begin{equation}
    \phi_{W}(x)=\oplus(f_{\theta_1}(x),f_{\theta_2}(x),...,f_{\theta_M}(x))
\end{equation}
A universal representation of the current concept is portrayed by an empirical mean of data samples of a mini-batch $B$.      
\begin{equation}
    \phi_{W}(X_B)=\frac{1}{|X_B|}\sum_{x\in X_B}\phi_{W}(x)
\end{equation}
The query $q(X_{B})$ of the current concept is defined as a weighted linear transformation of the query parameters $\{W_q,b_q\}$ and the universal representation $\phi_{W}(X_B)$ as follows:
\begin{equation}
q(X_{B})=W_{q}\phi_{W}(X_B)+b_q
\end{equation}
The keys $k_{i}(X_{B})$ are defined for all fast learners and rely on similar linear transformation as per the queries parameterized by $\{W_{k},b_{k}\}$.
\begin{equation}
    k_{i}(X_{B})=W_{k}f_{\theta_i}(X_{B})+b_k
\end{equation}
where $f_{\theta_i}(X_{B})=\frac{1}{|X_{B}|}\sum_{x \in X_{B}}f_{\theta_i}(x)$. Based on the query-key pairs, the attention score $\alpha_i$ is defined as per the original transformer \cite{Vaswani2017AttentionIA} as follows:
\begin{equation}\label{attentionscore}
    \alpha_i=\frac{\beta_i}{\sum_{i'=1}^{M}\beta_{i'}}; \beta_{i'}=\frac{q(X_{B})^{\top}k_i(X_B)}{\sqrt{l}}
\end{equation}
where $l$ is the dimension of the keys and queries. Since we deal with the forecasting problems here, there does not exist any aggregation necessary for the full support set. The output of the attention score is produced by mixing the attention score and the fast learner representations as follows:
\begin{equation}
    \phi_{W}(x)= \sum_{i=1}^{M}\alpha_i f_{\theta_i}(x)
\end{equation}
It is obvious that the URT mechanism is capable of selecting a particular fast learner with a maximum attention $\alpha_i=1$ or combining them with varying attention scores, i.e., the attention score satisfies the partition of unity $\sum_{i}\alpha_i=1$ \eqref{attentionscore}. 

MANTRA can be structured under a multi-head configuration receiving the same input as the first URT layer $\{f_{\theta_i}(.)\}_{i=1}^{M}$. The output of MANTRA under the multi-head configuration is expressed as follows:
\begin{equation}\label{multi-head}
    \phi_{W}(x)=\oplus(\{\phi_{W_j}(x)\}_{j=1}^{S})
\end{equation}
where $S$ denotes the number of heads in the URT structure. A regularization approach is implemented to prevent duplication of attention scores as per \cite{Liu2021AUR} $\Omega(W)=||AA^{\top} - I||_{F}^{2}$. A linear transformation $\{W_f,b_f\}$ is associated with \eqref{multi-head} by following the same concept as the original transformer \cite{Vaswani2017AttentionIA} producing the final output of MANTRA.
\begin{equation}
    \phi_{W}(x)=\sum_{j=1}^{S}\phi_{W_j}(x)W_{f}^{j}+b_{f}^{j}
\end{equation}
A fast adaptation can be achieved by only updating the most recent URT layer $\phi_{W_S}(x)$ with the query-key parameters $\{W_q^{j},b_q^{j}\}$ and $\{W_{k}^{j},b_k^{j}\}$ and the multi-head linear layer parameters $\{W_f,b_f\}$ leaving other parameters fixed. This strategy assures fast convergence to a new concept. Pseudo-codes of MANTRA are shown in Algorithm \ref{algorithm1}.

\begin{algorithm}
\caption{Learning Policy of MANTRA}\label{algorithm1}
\begin{algorithmic}[1]
\State \textbf{Input:} Fast learners $\{f_{\theta_{i}}(.)\}_{i=1}^{M}$, slow learner $g_{\psi}(.)$, URT layer $\phi_{W}(.)$, number of epochs $E$, number of batches of train dataset $B$.
\State \textbf{Output:}  Updated Networks Parameters $\{f_{\theta_{i}}(.)\}_{i=1}^{M}$, $g_{\psi}(.)$, $\phi_{W}(.)$
\item[\hspace{5 mm} // Train Fast and Slow Learners]
\For{$e=1:E$}
    \For{$b=1:B$}
        \State $\mathcal{B}$ = minibatch sample of train dataset
        %\State Predict output using averaged slow learners prediction    
        \State Compute loss $\mathcal{L}_{MSE}$
        \State Update fast learners parameters $\{f_{\theta_{i}}(.)\}_{i=1}^{M}$
        \State Compute $m_t$ and generate $\tilde{x}_t$ 
        \State Compute loss $\mathcal{L}_{S}$ following eq. (6)
        \State Update slow learner parameters $g_{\psi}(.)$ based on $\mathcal{L}_{S}$
        \State Update fast learners parameters $\{f_{\theta_{i}}(.)\}_{i=1}^{M}$ based on $\mathcal{L}_{S}+\mathcal{L}_{MSE}$
    \EndFor
\EndFor
\State Freeze fast learners parameters $\{f_{\theta_{i}}(.)\}_{i=1}^{M}$, 

\item[\hspace{5 mm} // Train URT Layer]
\For{$e=1:E$}
    \For{$b=1:B$}
        \State $\mathcal{B}$ = minibatch sample of train dataset
        %\State Predict output using Fast Learners and URT eq. (16)
        \State Compute loss $\mathcal{L}_{MSE}$
        \State Update URT parameters $\phi_{W}(.)$
    \EndFor
\EndFor
\end{algorithmic}
\end{algorithm}

%\subsection{Complexity Analysis}
\subsection{Complexity Analysis}
This sub-section discusses the complexity analysis of the proposed method. Suppose that  $N$ is the total samples of the training dataset that are divided into batches B that satisfies $\mathcal{\sum}_{b=1}^{B} N_b = N$, $E$ is the number of epochs for networks training, $M$ is the number of fast learners. Please note that the proposed method only uses 1 slow learner. Following MANTRA learning policy presented in Algorithm 1, there are two main processes i.e. train fast and slow learners and train the URT layer. Those processes contain atomic operation i.e. prediction, computing loss, and network update that are conducted with complexity $O(1)$ for one learner. Therefore, prediction and update by $M$ fast learners have $O(M)$ complexity.  Let $C$ denote the complexity of a process. Following the Algorithm 1, the complexity of the proposed method can be written as the following equations:

\begin{equation}
    \begin{split}
        \mathcal{C}(MANTRA) & =\mathcal{C}(train Fast and Slow) + \mathcal{C}(train URT)
    \end{split}
\end{equation}

\begin{equation}
    \begin{split}
        \mathcal{C}(MANTRA) & =E.\mathcal{\sum}_{b=1}^{B} N_b (O(M)+O(1)+O(M)+ \\ & O(1)+O(1)+O(1)+O(M)) \\
        & + E.\mathcal{\sum}_{b=1}^{B} N_b (O(M)+O(1)+O(1))+ 
    \end{split}
\end{equation}

\begin{equation}
    \begin{split}
        \mathcal{C}(MANTRA) & =E.\mathcal{\sum}_{b=1}^{B} N_b (O(M)) + E.\mathcal{\sum}_{b=1}^{B} N_b (O(M))
    \end{split}
\end{equation}

\begin{equation}
    \begin{split}
        \mathcal{C}(MANTRA) & = O(E.M.\mathcal{\sum}_{b=1}^{B} N_b)+  O(E.M.\mathcal{\sum}_{b=1}^{B} N_b)
    \end{split}
\end{equation}

Since $\mathcal{\sum}_{b=1}^{B} N_b = N$, then the compexity of MANTRA can be written as:

\begin{equation}
    \begin{split}
        \mathcal{C}(MANTRA) & = O(E.M.N) +  O(E.M.N)
    \end{split}
\end{equation}

\begin{equation}
    \begin{split}
        \mathcal{C}(MANTRA) & = O(E.M.N)
    \end{split}
\end{equation}

The complexity of the proposed method is $O(E.M.N)$ where $E$ is the training epoch, $M$ is the number of fast learners and $N$ is the size of the training dataset. In the case where the number of fast learners $M$ is a small constant $< 10$ and the number of epochs is also a constant, then the complexity of the proposed method is $O(N)$. Table \ref{complexity} reports the complexities of MANTRA compared to the ensemble of Autoformer and Autoformer. It is perceived that MANTRA's complexity is in the same order as the ensemble of Autoformer and Autoformer, $O(N)$. It is also seen that execution times per epoch for MANTRA are higher than those of Autoformer and ensemble of Autoformer in the ETT dataset because it involves additional training procedures, i.e., slow learning procedure and URT layer learning procedure, from those of Autoformer and ensemble of Autoformer. 

\begin{table}[h]
\centering
\begin{tabular}{cccccc}
\hline
\multirow{2}{*}{Method} & \multirow{2}{*}{Compelxity} & \multicolumn{4}{c}{Per epoch running time on ETT (s)} \\ \cline{3-6}
                        &                             & 96            & 192           & 336           & 720           \\ \hline
Autoformer              & $O(N)$                        & 47.48         & 57.9          & 74.04         & 123.44        \\
Ens.Autoformer          & $O(N)$                        & 91.83         & 101.92        & 118.45        & 188.54        \\
MANTRA(Ours)            & $O(N)$                        & 140.75        & 156.42        & 181.79        & 299.61  \\ \hline     
\end{tabular}
\label{complexity}
\caption{Complexities of Different Methods.}
\end{table}

\section{Experiments}
 This section discusses our numerical study on four real-world datasets in four different application domains: energy, economics, weather and disease. 

\subsection{Datasets}
The ETT dataset is applied to test the performance of consolidated algorithms in which it is formed by electricity transformers data covering load and oil temperature collected every 15 minutes over the period of July 2016 and July 2018. The weather dataset contains 2020 weather data collected every 10 minutes. It consists of 21 meteorological indicators, such as air temperature, humidity, etc. The ILI dataset encompasses weekly recorded influenza-like illness (ILI) patient data obtained from the Centers for Disease Control and Prevention of the United States over a period of 2002 and 2021. Particularly, it presents the ratio of patients seen with ILI and the total number of patients. The exchange dataset consists of the daily exchange rates of eight different countries over the period of 1990 to 2016. We follow \cite{Wu2021AutoformerDT} for the dataset split where the ratio of 6:2:2 is implemented for the ETT dataset while the ratio of 7:1:2 is used for other datasets. These problems are simulated in both univariate and multivariate forecasting problems. 

\subsection{Implementation Details}
We follow the same setting as \cite{Wu2021AutoformerDT} where $L_{2}$ loss is applied using the ADAM optimizer while the initial learning rate of $10^{-2}$ with an early stopping criterion is applied. The batch size is fixed to $32$. Our numerical results are produced using a single NVIDIA A5000 GPU. The fast learner and the slow learner are configured as 2 encoder layers and 1 decoder layer. As with \cite{Wu2021AutoformerDT}, the hyper-parameter $c$ of auto-correlation lies in the range of $1$ to $3$ while the number of URT heads and fast learners are set to $1$ and $3$ respectively. 

\subsection{Baselines}
MANTRA is compared with four state-of-the art transformer-based models: Autoformer \cite{Wu2021AutoformerDT}, Informer \cite{Zhou2021InformerBE}, Reformer \cite{Kitaev2020ReformerTE}, LogTrans \cite{LI2019EnhancingTL}, two recurrent-based model LSTNet \cite{Lai2018ModelingLA}, LSTM\cite{hochreiter1997long} and one convolutional-based model TCN \cite{Bai2018AnEE}. Comparisons are performed for both multivariate and univariate configurations. Autoformer is simulated in the same computational environments using its official implementations where its hyper-parameters are set using the grid search technique. Other numerical results are taken from \cite{Wu2021AutoformerDT}. Numerical reports are reported from an average of three independent executions under different random seeds.

\subsection{Numerical Results}
The advantage of MANTRA is demonstrated in Table \ref{tab:full_multivar} and \ref{tab:full_univar} for multivariate and univariate forecasting respectively. In the realm of multivariate forecasting, MANTRA outperforms other algorithms with noticeable margins across all predictive lengths. MANTRA is only on par with Autoformer in the predictive length of 96 of the exchange dataset. As the predictive length increases, MANTRA outperforms Autoformer with notable margins. On the other hand, the same finding is observed in the univariate forecasting setting where MANTRA beats other algorithms in most cases. MANTRA is only inferior to Informer in the predictive length of 96 of the ETT dataset and in the predictive length of 96 and 192 of the weather dataset. Although MANTRA is behind in the three cases, this fact can be caused by the base learner where MANTRA applies Autoformer as its base learner. Informer exceeds Autoformer in the three cases. Note that the concept of MANTRA is applicable to any base learners. Percentage improvements of MANTRA w.r.t. other algorithms are reported in Table \ref{tab:imp_multivar} and \ref{tab:imp_univar}. In addition, Table \ref{statistical_test} reports the t-test between MANTRA and Autoformer. It is seen that MANTRA outperforms Autoformer with statistically significant margins in most cases. 

\begin{figure}
    \centering
    \includegraphics[scale=0.2]{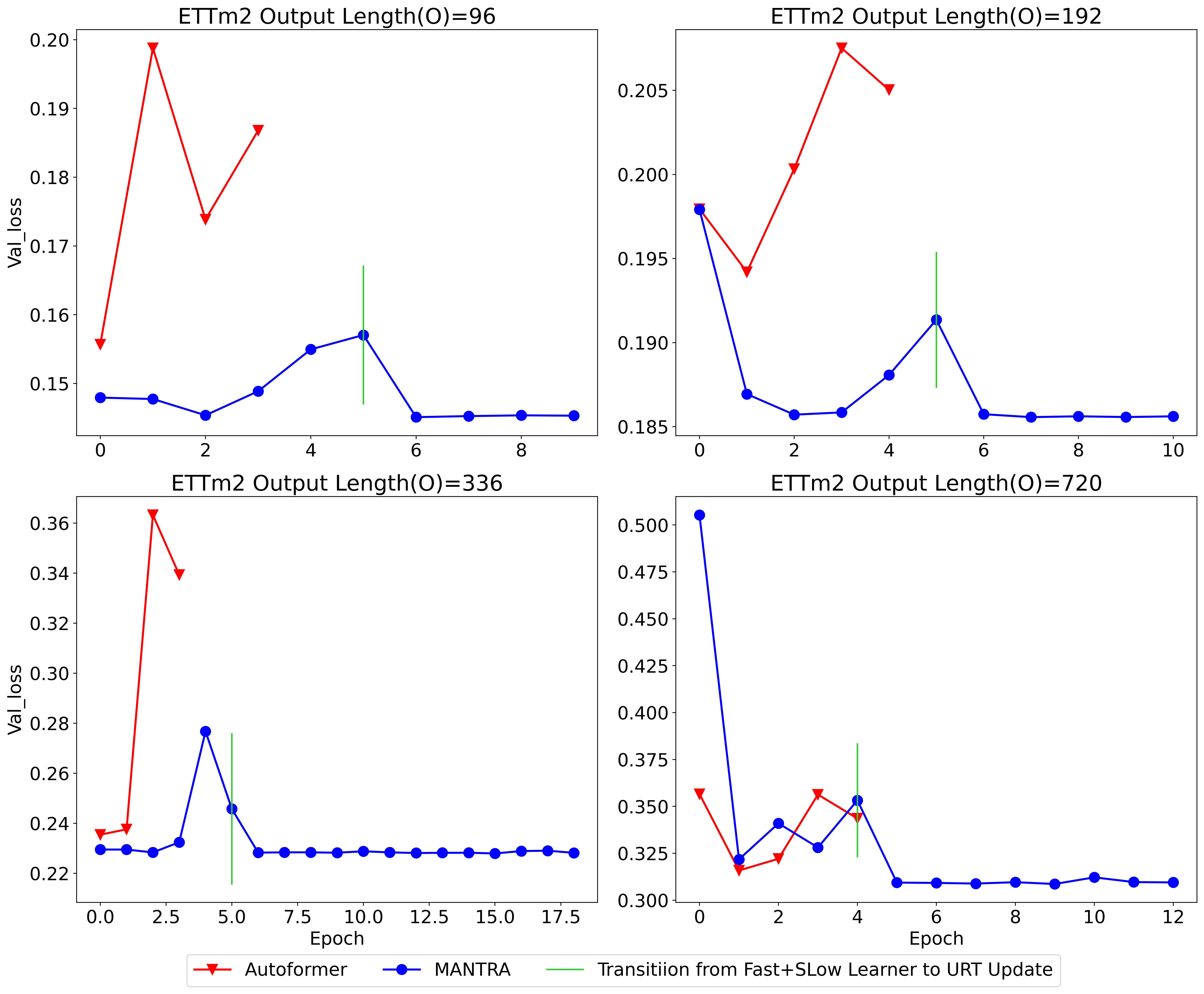}
    \caption{\textcolor{red}{MANTRA and Autoformer validation loss on ETTm dataset}}
    \label{fig:MANTRA_val_loss}
\end{figure}

\begin{figure}
    \centering
    \includegraphics[scale=0.2]{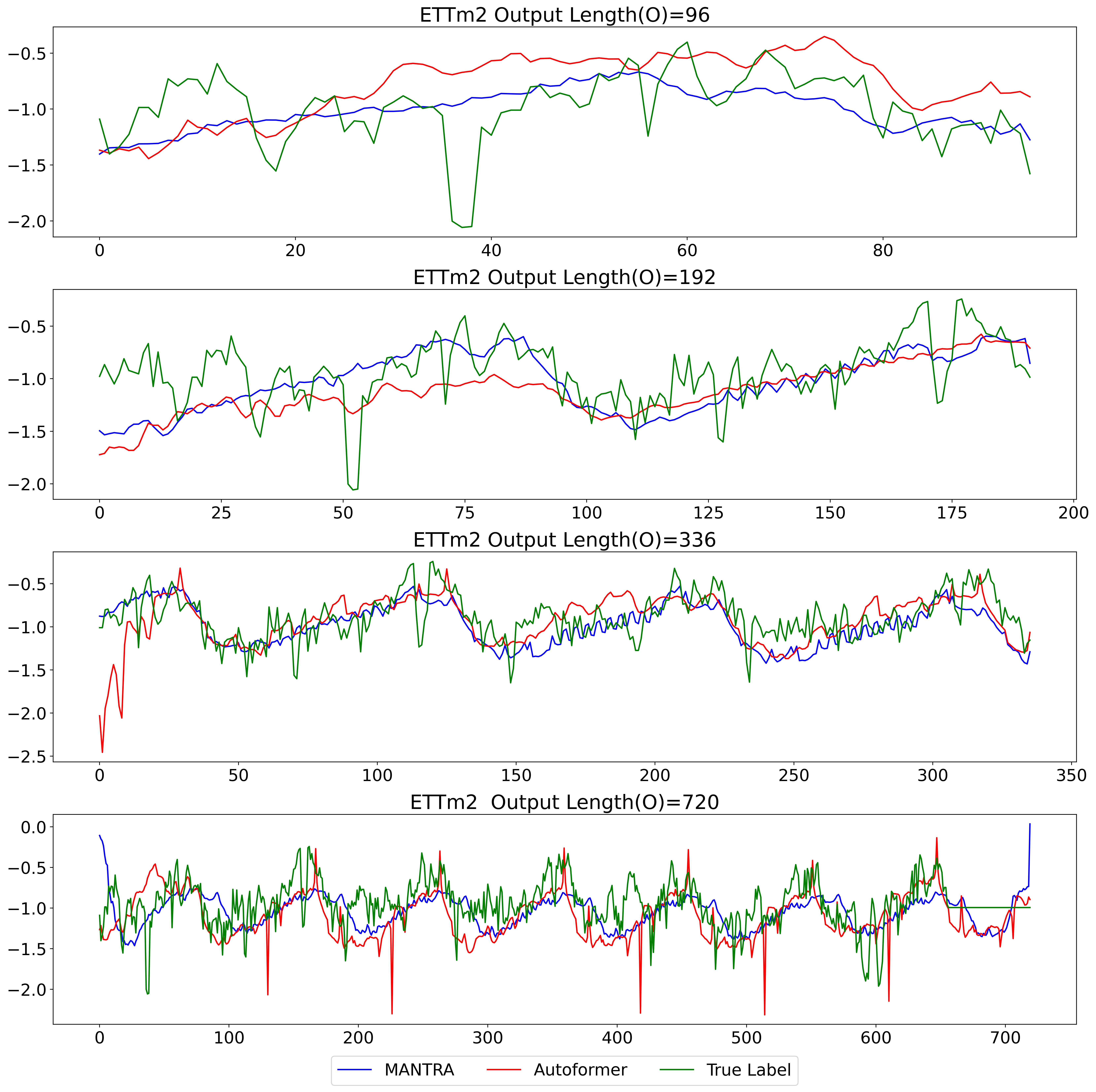}
    \caption{Visualization of MANTRA prediction compared to actual label on ETTm dataset}
    \label{fig:MANTRA_prediction}
\end{figure}

\textcolor{red}{Fig. \ref{fig:MANTRA_val_loss} exhibits the validation losses of MANTRA and Autoformer in the ETT dataset. It is perceived that the validation losses of MANTRA are smaller than Autoformer across all predictive lengths. One may wonder the increase of validation losses in Autoformer and MANTRA. This is caused by the fact that Autoformer adopts early stopping criteria and is trained with a few epochs. MANTRA's validation losses increase later than Autoformer. Importantly, the validation loss of MANTRA decreases further after switching to the URT update phase and fixing the base learners.} The advantage of MANTRA compared to Autoformer is also depicted in Fig. \ref{fig:MANTRA_prediction} in which it delivers more accurate predictions than Autoformer across all predictive lengths of the ETT dataset. 

\subsection{Ablation Studies}
Ablation studies are performed to study the influence of MANTRA's learning modules. Table \ref{tab:ablation} reports our numerical results produced with the ETT and ILI datasets. The URT layer plays a key role for MANTRA where its absence brings down numerical results significantly in both the ETT dataset and the ILI dataset. The URT layer allows fast adaptations of new patterns because base models can be frozen while only updating a few parameters of the URT layer. On the other hand, the slow learner contributes positively to the overall performance of MANTRA where its absence deteriorates the numerical results of MANTRA. In the ETT dataset, its effect is not obvious for small predictive lengths but becomes clear with the increase in predictive length. In the ILI dataset, MANTRA's performances are compromised due to the absence of the slow learner for all cases. This fact confirms the efficacy of the extended dual network concept which decouples the representation learning and the discriminative learning. The slow learner trained in a self-supervised manner is capable of generating useful representations of downstream tasks.  

\subsection{Sensitivity Analysis}
Sensitivity Analysis is performed to study the effect of the URT head and fast learners which it is done with the ETT dataset. Numerical results are reported in Table \ref{tab:sens_nlearner} and \ref{tab:sens_urthead} respectively. The increase of the fast learner improves the numerical results up to 3 fast learners but this trend does not continue with over 3 fast learners. This finding is mainly attributed to the random initialization strategy of MANTRA which does not generate sufficiently diverse fast learners. One might find improved strategies for fast learner creations. On the other hand, the increase of URT heads does not contribute toward the improvement of MANTRA's performances and even affects negatively the computational burdens.  

\begin{table*}[]
% \small
\centering
\begin{tabular}{c|c|cccccccccccccccc}
\hline
\multicolumn{2}{c}{{Models}}                              & \multicolumn{2}{c}{{MANTRA(Ours)}}           & \multicolumn{2}{c}{{Autoformer\cite{Wu2021AutoformerDT}}}       & \multicolumn{2}{c}{{Informer\cite{Zhou2021InformerBE}{}{}}} & \multicolumn{2}{c}{{LogTrans\cite{LI2019EnhancingTL}{}{}}} & \multicolumn{2}{c}{{Reformer\cite{Kitaev2020ReformerTE}{}{}}} & \multicolumn{2}{c}{{LSTNet\cite{Lai2018ModelingLA}{}{}}}  & \multicolumn{2}{c}{{LSTM\cite{hochreiter1997long}{}{}}}    & \multicolumn{2}{c}{{TCN\cite{Bai2018AnEE}{}{}}}      \\ \hline
\multicolumn{2}{c}{{Metrics}}                             & {MSE}   & {MAE}   & {MSE}   & {MAE}   & {MSE}    & {MAE}   & {MSE}    & {MAE}   & {MSE}    & {MAE}   & {MSE}   & {MAE}   & {MSE}   & {MAE}   & {MSE}   & {MAE}   \\ \hline
{}                           & {96}  & {\textbf{0.212}} & {\textbf{0.295}} & {0.255}          & {0.326}          & {0.365} & {0.453} & {0.768} & {0.642} & {0.658} & {0.619} & {3.142} & {1.365} & {2.041} & {1.073} & {3.041} & {1.330} \\
{}                           & {192} & {\textbf{0.275}} & {\textbf{0.335}} & {0.299}          & {0.350}          & {0.533} & {0.563} & {0.989} & {0.757} & {1.078} & {0.827} & {3.154} & {1.369} & {2.249} & {1.112} & {3.072} & {1.339} \\
{}                           & {336} & {\textbf{0.327}} & {\textbf{0.365}} & {0.337}          & {0.372}          & {1.363} & {0.887} & {1.334} & {0.872} & {1.549} & {0.972} & {3.160} & {1.369} & {2.568} & {1.238} & {3.105} & {1.348} \\
\multirow{-4}{*}{\rotatebox[origin=c]{90}{ETT}}      & {720} & {\textbf{0.440}} & {\textbf{0.435}} & {0.442}          & {\textbf{0.432}} & {3.379} & {1.388} & {3.048} & {1.328} & {2.631} & {1.242} & {3.171} & {1.368} & {2.720} & {1.287} & {3.135} & {1.354} \\ \hline
{}                           & {96}  & {\textbf{0.155}} & {\textbf{0.285}} & {\textbf{0.153}} & {\textbf{0.285}} & {0.847} & {0.752} & {0.968} & {0.812} & {1.065} & {0.829} & {1.551} & {1.058} & {1.453} & {1.049} & {3.004} & {1.432} \\
{}                           & {192} & {\textbf{0.266}} & {\textbf{0.377}} & {0.295}          & {0.395}          & {1.204} & {0.895} & {1.040} & {0.851} & {1.188} & {0.906} & {1.477} & {1.028} & {1.846} & {1.179} & {3.048} & {1.444} \\
{}                           & {336} & {\textbf{0.421}} & {\textbf{0.480}} & {0.446}          & {0.496}          & {1.672} & {1.036} & {1.659} & {1.081} & {1.357} & {0.976} & {1.507} & {1.031} & {2.136} & {1.231} & {3.113} & {1.459} \\
\multirow{-4}{*}{\rotatebox[origin=c]{90}{Exchange}} & {720} & {\textbf{1.168}} & {\textbf{0.847}} & {1.503}          & {0.919}          & {2.478} & {1.310} & {1.941} & {1.127} & {1.510} & {1.016} & {2.285} & {1.243} & {2.984} & {1.427} & {3.150} & {1.458} \\ \hline
{}                           & {96}  & {\textbf{0.248}} & {\textbf{0.321}} & {0.269}          & {0.338}          & {0.300} & {0.384} & {0.458} & {0.490} & {0.689} & {0.596} & {0.594} & {0.587} & {0.369} & {0.406} & {0.615} & {0.589} \\
{}                           & {192} & {\textbf{0.281}} & {\textbf{0.338}} & {0.297}          & {0.354}          & {0.598} & {0.544} & {0.658} & {0.589} & {0.752} & {0.638} & {0.560} & {0.565} & {0.416} & {0.435} & {0.629} & {0.600} \\
{}                           & {336} & {\textbf{0.329}} & {\textbf{0.369}} & {0.358}          & {0.392}          & {0.578} & {0.523} & {0.797} & {0.652} & {0.639} & {0.596} & {0.597} & {0.587} & {0.455} & {0.454} & {0.639} & {0.608} \\
\multirow{-4}{*}{\rotatebox[origin=c]{90}{Weather}}  & {720} & {\textbf{0.405}} & {\textbf{0.414}} & {0.450}          & {0.452}          & {1.059} & {0.741} & {0.869} & {0.675} & {1.130} & {0.792} & {0.618} & {0.599} & {0.535} & {0.520} & {0.639} & {0.610} \\ \hline
{}                           & {24}  & {\textbf{3.238}} & {\textbf{1.224}} & {3.680}          & {1.346}          & {5.764} & {1.677} & {4.480} & {1.444} & {4.400} & {1.382} & {6.026} & {1.770} & {5.914} & {1.734} & {6.624} & {1.830} \\
{}                           & {36}  & {\textbf{2.964}} & {\textbf{1.176}} & {3.329}          & {1.260}          & {4.755} & {1.467} & {4.799} & {1.467} & {4.783} & {1.448} & {5.340} & {1.668} & {6.631} & {1.845} & {6.858} & {1.879} \\
{}                           & {48}  & {\textbf{2.941}} & {\textbf{1.144}} & {3.376}          & {1.258}          & {4.763} & {1.469} & {4.800} & {1.468} & {4.832} & {1.465} & {6.080} & {1.787} & {6.736} & {1.857} & {6.968} & {1.892} \\
\multirow{-4}{*}{\rotatebox[origin=c]{90}{ILI}}      & {60}  & {\textbf{2.705}} & {\textbf{1.106}} & {2.917}          & {1.159}          & {5.264} & {1.564} & {5.278} & {1.560} & {4.882} & {1.483} & {5.548} & {1.720} & {6.870} & {1.879} & {7.127} & {1.918} \\ \hline
\end{tabular}
\caption{Multivariate results with different prediction lengths O E \{96, 192, 336, 720\}, averaged across 3 times run. We set the
input length I as 36 for ILI and 96 for the others. }
\label{tab:full_multivar}
\end{table*}

\begin{table*}[]
\centering
\begin{tabular}{c|c|cccccccccc}
\hline
\multicolumn{2}{c}{{Models}}                              & \multicolumn{2}{c}{{\textbf{MANTRA(Ours)}}}                      & \multicolumn{2}{c}{{Autoformer\cite{Wu2021AutoformerDT}}}         & \multicolumn{2}{c}{{Informer\cite{Zhou2021InformerBE}}}                             & \multicolumn{2}{c}{{LogTrans\cite{LI2019EnhancingTL}}}           & \multicolumn{2}{c}{{Reformer\cite{Kitaev2020ReformerTE}}}           \\ \hline
\multicolumn{2}{c}{{Mettrics}}                            & {MSE}             & {MAE}             & {MSE}    & {MAE}    & {MSE}             & {MAE}             & {MSE}    & {MAE}    & {MSE}    & {MAE}    \\ \hline
{}                           & {96}  & {0.1035}          & {0.2447}          & {0.1043} & {0.2455} & {\textbf{0.0706}} & {\textbf{0.2030}} & {0.0824} & {0.2309} & {0.1740} & {0.3424} \\
{}                           & {192} & {\textbf{0.1344}} & {\textbf{0.2844}} & {0.1574} & {0.3042} & {0.2093}          & {0.3797}          & {0.1709} & {0.3296} & {0.2409} & {0.4067} \\
{}                           & {336} & {\textbf{0.1631}} & {\textbf{0.3145}} & {0.1608} & {0.3104} & {0.3041}          & {0.4743}          & {0.2822} & {0.4466} & {0.3538} & {0.5153} \\
\multirow{-4}{*}{{ETT}}      & {720} & {\textbf{0.1853}} & {\textbf{0.3359}} & {0.2007} & {0.3500} & {0.3571}          & {0.5110}          & {0.4313} & {0.5817} & {0.3759} & {0.5417} \\ \hline
{}                           & {96}  & {\textbf{0.1427}} & {\textbf{0.2970}} & {0.1639} & {0.3179} & {0.5172}          & {0.5694}          & {0.2737} & {0.3999} & {0.5757} & {0.6101} \\
{}                           & {192} & {\textbf{0.2815}} & {\textbf{0.4153}} & {0.3278} & {0.4413} & {1.0956}          & {0.8252}          & {1.7398} & {0.9529} & {0.9056} & {0.7751} \\
{}                           & {336} & {\textbf{0.5099}} & {\textbf{0.5522}} & {0.6178} & {0.6097} & {1.4225}          & {0.9660}          & {2.4667} & {1.1975} & {2.0816} & {1.2192} \\
\multirow{-4}{*}{{Exchange}} & {720} & {\textbf{1.2448}} & {\textbf{0.8677}} & {1.3049} & {0.8960} & {1.3918}          & {0.9332}          & {2.3174} & {1.1825} & {2.3128} & {1.3229} \\ \hline
{}                           & {96}  & {0.0079}          & {0.0705}          & {0.0159} & {0.1001} & {\textbf{0.0043}} & {\textbf{0.0478}} & {0.0041} & {0.0511} & {0.0046} & {0.0542} \\
{}                           & {192} & {0.0062}          & {0.0582}          & {0.0086} & {0.0688} & {\textbf{0.0046}} & {\textbf{0.0521}} & {0.0048} & {0.0499} & {0.0433} & {0.0927} \\
{}                           & {336} & {\textbf{0.0040}} & {\textbf{0.0475}} & {0.0050} & {0.0543} & {0.0044}          & {0.0508}          & {0.0088} & {0.0721} & {0.0149} & {0.1019} \\
\multirow{-4}{*}{{Weather}}  & {720} & {\textbf{0.0040}} & {\textbf{0.0480}} & {0.0083} & {0.0707} & {0.0050}          & {0.0535}          & {0.0064} & {0.0628} & {0.0195} & {0.1168} \\ \hline
{}                           & {24}  & {\textbf{0.7315}} & {\textbf{0.6411}} & {0.7948} & {0.6827} & {5.6955}          & {2.1385}          & {3.9573} & {1.7524} & {3.7231} & {1.7053} \\
{}                           & {36}  & {\textbf{0.6986}} & {\textbf{0.6366}} & {0.7157} & {0.6848} & {5.1962}          & {2.0467}          & {3.9925} & {1.7635} & {3.6082} & {1.6725} \\
{}                           & {48}  & {\textbf{0.7427}} & {\textbf{0.7054}} & {0.8019} & {0.7482} & {4.8515}          & {1.9779}          & {3.9118} & {1.7542} & {3.8564} & {1.7636} \\
\multirow{-4}{*}{{ILI}}      & {60}  & {\textbf{0.8327}} & {\textbf{0.7645}} & {0.9219} & {0.8205} & {5.3332}          & {2.0800}          & {3.9626} & {1.7741} & {4.3197} & {1.8932} \\ \hline
\end{tabular}
\caption{Univariate results with different prediction lengths O E \{96, 192, 336, 720\}, averaged across 3 times run. We set the
input length I as 36 for ILI and 96 for the others. }
\label{tab:full_univar}
\end{table*}

\begin{table*}[]
\centering
\begin{tabular}{c|c|cccccccccccccc}
\hline
\multicolumn{2}{c}{}            & \multicolumn{14}{c}{MANTRA Improvement (\%) compared to the competitor methods}                                                                                                                                                                              \\ \hline
\multicolumn{2}{c}{Models}      & \multicolumn{2}{c}{Autoformer\cite{Wu2021AutoformerDT}} & \multicolumn{2}{c}{Informer\cite{Zhou2021InformerBE}{}{}} & \multicolumn{2}{c}{LogTrans\cite{LI2019EnhancingTL}{}{}} & \multicolumn{2}{c}{Reformer\cite{Kitaev2020ReformerTE}{}{}} & \multicolumn{2}{c}{LSTNet\cite{Lai2018ModelingLA}{}{}} & \multicolumn{2}{c}{LSTM\cite{hochreiter1997long}{}{}} & \multicolumn{2}{c}{TCN\cite{Bai2018AnEE}{}{}} \\ \hline
\multicolumn{2}{c}{Metrics}     & MSE            & MAE           & MSE               & MAE              & MSE               & MAE              & MSE               & MAE              & MSE              & MAE             & MSE             & MAE            & MSE            & MAE           \\ \hline
\multirow{4}{*}{ETT}      & 96  & 16.63          & 9.57          & 41.86             & 34.85            & 72.37             & 54.03            & 67.75             & 52.32            & 93.25            & 78.38           & 89.60           & 72.50          & 93.02          & 77.81         \\
                          & 192 & 7.99           & 4.31          & 48.34             & 40.47            & 72.16             & 55.73            & 74.46             & 59.47            & 91.27            & 75.52           & 87.76           & 69.86          & 91.04          & 74.97         \\
                          & 336 & 2.87           & 1.90          & 76.02             & 58.88            & 75.50             & 58.18            & 78.90             & 62.48            & 89.66            & 73.36           & 87.27           & 70.54          & 89.47          & 72.94         \\
                          & 720 & 0.41           & -0.70         & 86.97             & 68.69            & 85.55             & 67.28            & 83.26             & 65.01            & 86.11            & 68.23           & 83.81           & 66.24          & 85.95          & 67.91         \\ \hline
\multirow{4}{*}{Exchange} & 96  & -0.74          & -0.11         & 81.75             & 62.08            & 84.03             & 64.89            & 85.48             & 65.61            & 90.03            & 73.05           & 89.36           & 72.82          & 94.85          & 80.09         \\ 
                          & 192 & 9.71           & 4.43          & 77.88             & 57.84            & 74.39             & 55.66            & 77.58             & 58.35            & 81.97            & 63.29           & 85.57           & 67.99          & 91.26          & 73.87         \\
                          & 336 & 5.67           & 3.14          & 74.81             & 53.64            & 74.62             & 55.57            & 68.97             & 50.79            & 72.06            & 53.41           & 80.29           & 60.98          & 86.47          & 67.08         \\
                          & 720 & 22.30          & 7.83          & 52.88             & 35.34            & 39.84             & 24.84            & 22.67             & 16.62            & 48.90            & 31.85           & 60.87           & 40.64          & 62.93          & 41.90         \\ \hline
\multirow{4}{*}{Weather}  & 96  & 7.67           & 5.01          & 17.17             & 16.38            & 45.74             & 34.47            & 63.93             & 46.12            & 58.17            & 45.30           & 32.66           & 20.91          & 59.60          & 45.48         \\
                          & 192 & 5.34           & 4.54          & 53.01             & 37.79            & 57.29             & 42.55            & 62.63             & 46.96            & 49.82            & 40.11           & 32.45           & 22.21          & 55.33          & 43.60         \\
                          & 336 & 7.93           & 5.76          & 43.05             & 29.42            & 58.70             & 43.38            & 48.49             & 38.06            & 44.87            & 37.11           & 27.66           & 18.69          & 48.49          & 39.29         \\
                          & 720 & 9.93           & 8.33          & 61.72             & 44.07            & 53.35             & 38.60            & 64.12             & 47.67            & 34.40            & 30.81           & 24.22           & 20.30          & 36.55          & 32.06         \\ \hline
\multirow{4}{*}{ILI}      & 24  & 12.01          & 9.05          & 43.83             & 27.02            & 27.73             & 15.24            & 26.42             & 11.44            & 46.27            & 30.85           & 45.25           & 29.42          & 51.12          & 33.12         \\
                          & 36  & 10.95          & 6.63          & 37.66             & 19.83            & 38.23             & 19.83            & 38.02             & 18.78            & 44.49            & 29.49           & 55.29           & 36.26          & 56.77          & 37.41         \\
                          & 48  & 12.90          & 9.04          & 38.26             & 22.10            & 38.74             & 22.04            & 39.14             & 21.88            & 51.63            & 35.96           & 56.34           & 38.37          & 57.80          & 39.51         \\
                          & 60  & 7.24           & 4.53          & 48.61             & 29.26            & 48.74             & 29.08            & 44.59             & 25.40            & 51.24            & 35.68           & 60.62           & 41.12          & 62.04          & 42.32   \\ \hline     
\end{tabular}
\caption{Improvement of MANTRA to competitor methods on multivariate setting with different prediction lengths  $O \in \{96,\, 192,\, 336,\, 720\}$, except for ILI dataset $O\in \{24,\, 36,\, 48,\, 60\}$}  
\label{tab:imp_multivar}
\end{table*}

\begin{table*}[]
\centering
\begin{tabular}{c|c|cccccccc}
 \hline  
\multicolumn{2}{c}{}            & \multicolumn{8}{c}{MANTRA Improvement (\%) compared to the competitor methods}                                              \\  \hline  
\multicolumn{2}{c}{Models}      & \multicolumn{2}{c}{Autoformer\cite{Wu2021AutoformerDT}} & \multicolumn{2}{c}{Informer\cite{Zhou2021InformerBE}} & \multicolumn{2}{c}{LogTrans\cite{LI2019EnhancingTL}} & \multicolumn{2}{c}{Reformer\cite{Kitaev2020ReformerTE}} \\  \hline  
\multicolumn{2}{c}{Mettrics}    & MSE            & MAE           & MSE           & MAE          & MSE           & MAE          & MSE           & MAE          \\  \hline  
\multirow{4}{*}{ETT}      & 96  & 0.75           & 0.36          & -46.63        & -20.49       & -25.63        & -5.96        & 40.49         & 28.54        \\
                          & 192 & 14.61          & 6.53          & 35.80         & 25.10        & 21.36         & 13.72        & 44.22         & 30.08        \\
                          & 336 & -1.48          & -1.33         & 46.35         & 33.69        & 42.18         & 29.59        & 53.89         & 38.97        \\
                          & 720 & 7.67           & 4.03          & 48.11         & 34.27        & 57.03         & 42.25        & 50.70         & 37.99        \\  \hline  
\multirow{4}{*}{Exchange} & 96  & 12.88          & 6.57          & 72.40         & 47.84        & 47.84         & 25.73        & 75.20         & 51.32        \\
                          & 192 & 14.15          & 5.90          & 74.31         & 49.68        & 83.82         & 56.42        & 68.92         & 46.42        \\
                          & 336 & 17.47          & 9.43          & 64.16         & 42.84        & 79.33         & 53.89        & 75.51         & 54.71        \\
                          & 720 & 4.60           & 3.16          & 10.57         & 7.02         & 46.29         & 26.62        & 46.18         & 34.41        \\  \hline  
\multirow{4}{*}{Weather}  & 96  & 50.41          & 29.51         & -82.45        & -47.67       & -91.70        & -38.09       & -72.84        & -30.14       \\
                          & 192 & 28.86          & 15.40         & -34.63        & -11.73       & -28.13        & -16.53       & 85.80         & 37.22        \\
                          & 336 & 21.17          & 12.53         & 9.97          & 6.50         & 54.94         & 34.09        & 73.46         & 53.38        \\
                          & 720 & 52.16          & 32.13         & 20.81         & 10.29        & 38.40         & 23.48        & 79.76         & 58.90        \\  \hline  
\multirow{4}{*}{ILI}      & 24  & 7.96           & 6.10          & 87.16         & 70.02        & 81.51         & 63.42        & 80.35         & 62.41        \\
                          & 36  & 2.39           & 7.04          & 86.56         & 68.90        & 82.50         & 63.90        & 80.64         & 61.94        \\
                          & 48  & 7.38           & 5.73          & 84.69         & 64.34        & 81.01         & 59.79        & 80.74         & 60.01        \\
                          & 60  & 9.68           & 6.83          & 84.39         & 63.25        & 78.99         & 56.91        & 80.72         & 59.62  \\  \hline       
\end{tabular}
\caption{Improvement of MANTRA to competitor methods on univariate setting with different prediction lengths  $O \in \{96,\, 192,\, 336,\, 720\}$, except for ILI dataset $O\in \{24,\, 36,\, 48,\, 60\}$ } 
\label{tab:imp_univar}
\end{table*}

\newcommand{\cmark}{\ding{51}}
\newcommand{\xmark}{\ding{55}}
\begin{table*}[]
\centering
\begin{tabular}{cccccccccc}
\hline
\multicolumn{2}{c}{}            & \multicolumn{4}{c}{Multivariate}                  & \multicolumn{4}{c}{Univariate}                    \\ \hline
\multicolumn{2}{c}{Mettrics}    & \multicolumn{2}{c}{MSE} & \multicolumn{2}{c}{MAE} & \multicolumn{2}{c}{MSE} & \multicolumn{2}{c}{MAE} \\ \hline
\multicolumn{2}{c}{T-Test}      & p-val        & Sig.     & P-val        & Sig.     & p-val        & Sig.     & P-val        & Sig.     \\ \hline
\multirow{4}{*}{ETT}      & 96  & 0.000662     & {\cmark}        & 0.000027     & {\cmark}        & 0.466713     & {\xmark}        & 0.459712     & {\xmark}        \\
                          & 192 & 0.002148     & {\cmark}        & 0.017119     & {\cmark}        & 0.029832     & {\cmark}        & 0.035679     & {\cmark}        \\
                          & 336 & 0.000393     & {\cmark}        & 0.003745     & {\cmark}        & 0.409776     & {\xmark}        & 0.323542     & {\xmark}        \\
                          & 720 & 0.456497     & {\xmark}        & 0.397439     & {\xmark}        & 0.082046     & {\cmark}        & 0.114526     & {\xmark}        \\ \hline 
\multirow{4}{*}{Exchange} & 96  & 0.457713     & {\xmark}        & 0.484059     & {\xmark}        & 0.029386     & {\cmark}        & 0.053399     & {\cmark}        \\
                          & 192 & 0.024725     & {\cmark}        & 0.029971     & {\cmark}        & 0.047900     & {\cmark}        & 0.053399     & {\cmark}        \\
                          & 336 & 0.101353     & {\xmark}        & 0.110363     & {\xmark}        & 0.096577     & {\cmark}        & 0.080948     & {\cmark}        \\
                          & 720 & 0.231046     & {\xmark}        & 0.268099     & {\xmark}        & 0.277424     & {\xmark}        & 0.225318     & {\xmark}        \\ \hline
\multirow{4}{*}{Weather}  & 96  & 0.085522     & {\cmark}        & 0.033795     & {\cmark}        & 0.004570     & {\cmark}        & 0.009602     & {\cmark}        \\
                          & 192 & 0.066209     & {\cmark}        & 0.057598     & {\cmark}        & 0.096289     & {\cmark}        & 0.035484     & {\cmark}        \\ 
                          & 336 & 0.021258     & {\cmark}        & 0.052073     & {\cmark}        & 0.002438     & {\cmark}        & 0.005528     & {\cmark}        \\
                          & 720 & 0.053112     & {\cmark}        & 0.035405     & {\cmark}        & 0.004743     & {\cmark}        & 0.006344     & {\cmark}        \\ \hline
\multirow{4}{*}{ILI}      & 24  & 0.069567     & {\cmark}        & 0.040147     & {\cmark}        & 0.000956     & {\cmark}        & 0.00531      & {\cmark}        \\
                          & 36  & 0.002188     & {\cmark}        & 0.007061     & {\cmark}        & 0.256694     & {\xmark}        & 0.000204     & {\cmark}        \\
                          & 48  & 0.007818     & {\cmark}        & 0.012957     & {\cmark}        & 0.078750     & {\cmark}        & 0.022501     & {\cmark}        \\ 
                          & 60  & 0.000010     & {\cmark}        & 0.006843     & {\cmark}        & 0.003887     & {\cmark}        & 0.003742     & {\cmark}       \\ \hline
\end{tabular}
\caption{T-Test with 0.1 significance level of MANTRA vs. Autoformer on multivariate and univariate setting} 
\label{statistical_test}
\end{table*}

\begin{table*}[]
\centering
\begin{tabular}{l|c|cccccccc}
\hline
{Config.} & {}                       & {MSE}    & {MAE}    & {MSE}    & {MAE}    & {MSE}    & {MAE}    & {MSE}    & {MAE}    \\ \hline
{}                            & {}                       & \multicolumn{2}{c}{{96}}                 & \multicolumn{2}{c}{{192}}                & \multicolumn{2}{c}{{336}}                & \multicolumn{2}{c}{{720}}                \\ \hline
{MANTRA}                      & {}                       & {0.2127} & {0.2951} & {0.2703} & {0.3296} & {0.3255} & {0.3625} & {0.4354} & {0.4286} \\
{MANTRA w/o URT}              & {}                       & {0.2305} & {0.3095} & {0.2850} & {0.3391} & {0.3387} & {0.3702} & {0.4340} & {0.4202} \\
{MANTRA w/o Slow Learner}     & \multirow{-3}{*}{{ETT}} & {0.2127} & {0.2934} & {0.2677} & {0.3249} & {0.3310} & {0.3682} & {0.4354} & {0.4304} \\ \hline
{}                            & {}                       & \multicolumn{2}{c}{{24}}                 & \multicolumn{2}{c}{{36}}                 & \multicolumn{2}{c}{{48}}                 & \multicolumn{2}{c}{{60}}                 \\ \hline
{MANTRA}                      & {}                       & {2.7930} & {1.1274} & {3.0280} & {1.1929} & {3.0008} & {1.1714} & {2.6418} & {1.0855} \\
{MANTRA w/o URT}              & {}                       & {5.2348} & {1.7377} & {4.7750} & {1.6171} & {4.3002} & {1.5188} & {4.3411} & {1.5185} \\
{MANTRA w/o Slow Learner}     & \multirow{-3}{*}{{ILI}}  & {3.6409} & {1.3103} & {3.4154} & {1.2342} & {3.3005} & {1.2076} & {3.1309} & {1.1821} \\ \hline
\end{tabular}
\caption{Ablation study on ETT and ILI dataset. }
\label{tab:ablation}
\end{table*}

\begin{table*}[]
\centering
\begin{tabular}{ccccccccc}
\hline
{}                            & \multicolumn{2}{c}{{96}}                                   & \multicolumn{2}{c}{{192}}                                  & \multicolumn{2}{c}{{336}}                                  & \multicolumn{2}{c}{{720}}                                  \\ \cline{2-9}
\multirow{-2}{*}{{N-Learner}} & {MSE}             & {MAE}             & {MSE}             & {MAE}             & {MSE}             & {MAE}             & {MSE}             & {MAE}             \\ \hline
{1}                           & {\textbf{0.2111}} & {\textbf{0.2936}} & {0.2706}          & {0.3275}          & {0.3323}          & {0.3718}          & {\textbf{0.4233}} & {\textbf{0.4177}} \\ 
{2}                           & {0.2114}          & {0.2925}          & {0.2828}          & {0.3425}          & {0.3262}          & {0.3637}          & {0.4261}          & {0.4237}          \\
{3}                           & {0.2127}          & {0.2951}          & {\textbf{0.2703}} & {\textbf{0.3296}} & {\textbf{0.3255}} & {\textbf{0.3625}} & {0.4354}          & {0.4286}          \\
{4}                           & {0.2167}          & {0.2997}          & {0.2738}          & {0.3311}          & {0.3413}          & {0.3789}          & {0.4298}          & {0.4231}          \\
{5}                           & {0.2157}          & {0.2992}          & {0.2721}          & {0.3312}          & {0.3311}          & {0.3687}          & {0.4341}          & {0.4271}          \\
{6}                           & {0.2159}          & {0.2995}          & {0.2723}          & {0.3309}          & {0.3376}          & {0.3783}          & {0.4270}          & {0.4242}          \\
{7}                           & {0.2125}          & {0.2956}          & {0.2726}          & {0.3310}          & {0.3347}          & {0.3715}          & {0.4473}          & {0.4434} \\ \hline         
\end{tabular}
\caption{Sensitivity analysis on ETT dataset with the different number of fast learners. }
\label{tab:sens_nlearner}
\end{table*}

\begin{table*}[]
\centering
\begin{tabular}{ccccccccc}
\hline
{}                           & \multicolumn{2}{c}{{96}}                 & \multicolumn{2}{c}{{192}}                & \multicolumn{2}{c}{{336}}                & \multicolumn{2}{c}{{720}}                \\ \cline{2-9}
\multirow{-2}{*}{{URT-Head}} & {MSE}    & {MAE}    & {MSE}    & {MAE}    & {MSE}    & {MAE}    & {MSE}    & {MAE}    \\ \hline
{1}                          & {0.2127} & {0.2951} & {0.2703} & {0.3296} & {0.3255} & {0.3625} & {0.4354} & {0.4286} \\
{2}                          & {0.2128} & {0.2952} & {0.2703} & {0.3297} & {0.3255} & {0.3625} & {0.4354} & {0.4286} \\
{3}                          & {0.2127} & {0.2952} & {0.2702} & {0.3296} & {0.3256} & {0.3625} & {0.4358} & {0.4292} \\
{4}                          & {0.2127} & {0.2952} & {0.2704} & {0.3298} & {0.3255} & {0.3625} & {0.4355} & {0.4287} \\
{5}                          & {0.2126} & {0.2951} & {0.2703} & {0.3296} & {0.3256} & {0.3626} & {0.4358} & {0.4291} \\
{6}                          & {0.2126} & {0.2951} & {0.2704} & {0.3297} & {0.3258} & {0.3627} & {0.4357} & {0.4290} \\
{7}                          & {0.2126} & {0.2951} & {0.2703} & {0.3296} & {0.3256} & {0.3626} & {0.4355} & {0.4286} \\ \hline
\end{tabular}
\caption{Sensitivity analysis on ETT dataset with the different number of URT heads. }
\label{tab:sens_urthead}
\end{table*}

\section{Conclusion}
This paper proposes the meta-transformer network (MANTRA) for dynamic long-term time-series forecasting tasks. MANTRA is developed from the concept of extended dual networks where an array of fast learners and a slow learner learn cooperatively with dual objectives minimizing predictive losses and self-supervised losses. The concept of URT is integrated to adapt quickly to concept drifts involving only a few parameters. That is, the URT layer produces concept-adapted representations enabling fast adaptations of concept changes. The advantage of MANTRA is numerically validated with four datasets and different predictive lengths under both multivariate and univariate settings. MANTRA outperforms other popular algorithms in 29 cases out of 32 cases with at least $3\%$ gaps. Although MANTRA is exemplified with Autoformer as a base learner here, it can be generalized to other base learners as well. Our future work is devoted to study the problem of time-series domain adaptations. 

\bibliographystyle{IEEEtran}
\bibliography{references}

\end{document}